%% file: iclr2027_conference.tex
\documentclass{article} %
\PassOptionsToPackage{table}{xcolor}
\usepackage{iclr2027_conference,times}

\input{math_commands.tex}

\usepackage{hyperref}
\usepackage{url}
\usepackage{enumitem}
\usepackage{graphicx}
\usepackage{booktabs}
\usepackage{capt-of,tabularx}
\usepackage{multirow}
\usepackage{pifont}
\usepackage[table]{xcolor}
\usepackage{titlesec}
\titlespacing*{\paragraph}{0pt}{0pt}{1em}
\definecolor{projectblue}{HTML}{246A9B}

\title{What 30,000 Hours of Ego-centric Video Does Not Teach}

\author{%
\textbf{Jiahua Dong$^{1}$\thanks{Work done during an internship at Toyota Research Institute.} \quad Anurag Bagchi$^{2}$ \quad Yash Jangir$^{3}$ \quad Muhammad Zubair Irshad$^{4}$}\\[2pt]
\textbf{Sergey Zakharov$^{4}$ \quad Martial Hebert$^{2}$ \quad Homanga Bharadhwaj$^{3}$}\\[2pt]
\textbf{Yu-Xiong Wang$^{1}$ \quad Vitor Campagnolo Guizilini$^{4}$ \quad Pavel Tokmakov$^{4}$}\\[6pt]
{\normalfont $^{1}$University of Illinois Urbana-Champaign \quad $^{2}$Carnegie Mellon University}\\
{\normalfont $^{3}$Johns Hopkins University \quad $^{4}$Toyota Research Institute}\\[12pt]
\multicolumn{1}{@{}c@{}}{\makebox[\textwidth][c]{%
  \hypersetup{hidelinks}%
  \normalfont\textcolor{projectblue}{%
    \url{https://fidelity-gap.github.io/}}%
}}
}

\hypersetup{
  pdftitle={What 30,000 Hours of Ego-centric Video Does Not Teach},
  pdfauthor={Jiahua Dong, Anurag Bagchi, Yash Jangir, Muhammad Zubair Irshad, Sergey Zakharov, Martial Hebert, Homanga Bharadhwaj, Yu-Xiong Wang, Vitor Campagnolo Guizilini, Pavel Tokmakov}
}

\iclrfinalcopy
\begin{document}

\maketitle
\vspace{-32pt}
\lhead{Preprint}
\input{sections/0_abstract}

\input{sections/1_intro}

\input{sections/2_related}

\input{sections/4_protocol}

\input{sections/5_experiment}

\input{sections/6_conclusion}

\subsection*{Acknowledgments}
We thank Mecka\footnote{\url{https://mecka.ai}} for contributing the ego-centric human video data used in this work.

\subsection*{AI use statement}
In this work, we used generative AI tools for interpreting results. We have not used generative AI tools for other tasks that require disclosure.   Additionally, we used generative AI tools for several tasks with recommended disclosure: editing parts of the paper, creating and modifying figures, editing software code, searching for related literature. We have reviewed all AI-assisted work. Every reported number is computed by scripted evaluation from saved model predictions and all characterizations of prior work were checked against the primary sources. We take responsibility for the final content of this work, including text or artifacts produced with the aid of generative AI.

\subsection*{Reproducibility statement}
Section~\ref{sec:preliminaries} describes the backbone, the two model variants as well as the skeleton conditioning, including the projection and colour encoding, and Section~\ref{sec:object_ceiling}
the dynamic-region localization and noise-scheduling objective, with the full algorithm and hyperparameters in Appendix~\ref{app:dynamic_region_details}. Section~\ref{sec:trainset}
describes the training corpus and the nested data ladder, with the sampling convention in Appendix~\ref{app:data_scaling_ladder}. Section~\ref{sec:eval} describes the evaluation benchmark, its curation criteria and mask annotation procedure, as well as the metrics, including our estimate of the
maximum score each category can attain. Section~\ref{sec:impl} lists optimization settings, and further training and implementation details. Appendix~\ref{app:humanoid} gives the humanoid fine-tuning setup and per-task results, and Appendix~\ref{app:drop} the skeleton-dropout control. We will publicly release our code, the pre-trained model checkpoints, and the evaluation benchmark with its instance mask annotations, so that every number we report can be reproduced.

\bibliography{iclr2027_conference}
\bibliographystyle{iclr2027_conference}

\appendix
\input{sections/a_appendix}
\input{sections/g_appendix}

\input{sections/c_appendix}
\input{sections/d_appendix}
\input{sections/b_appendix}
\input{sections/f_appendix}

\input{sections/e_appendix}

\end{document}

%% file: math_commands.tex
\usepackage{amsmath,amsfonts,bm}

\def\eqref#1{equation~\ref{#1}}

\def\1{\bm{1}}

\DeclareMathAlphabet{\mathsfit}{\encodingdefault}{\sfdefault}{m}{sl}
\SetMathAlphabet{\mathsfit}{bold}{\encodingdefault}{\sfdefault}{bx}{n}

%% file: sections/0_abstract.tex
\begin{figure}[h]
\centering
\includegraphics[width=\textwidth,trim=0 0 0 20bp,clip]{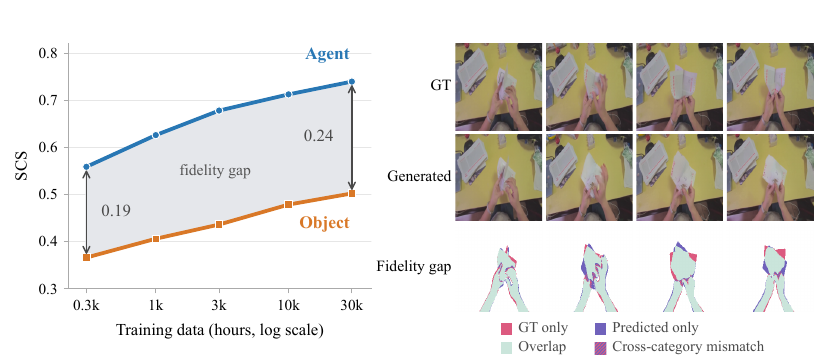}
\caption{\textbf{Limitations of world model scaling.} \emph{Left:} Ego-centric data scaling can achieve high agent modeling fidelity, but the effect of agent's actions on the world lags behind. \emph{Right:} The same failure at the level of a single prediction. Our best model places the hands almost exactly where they belong, while the paper they are folding is rendered in the wrong configuration throughout. See the \href{https://fidelity-gap.github.io/}{project website} for more video-level results.
}
\label{fig:agent_object_gap}
\end{figure}

\begin{abstract}
World models offer a promising alternative to physics-based simulators, yet remain far from practical deployment. We ask how far scaling ego-centric human video takes them, using a dataset of 30,000 hours spanning over 1,000 scene types and 14,000 contributors. Rather than relying on opaque downstream metrics, we directly evaluate agent and object-interaction fidelity on a challenging out-of-distribution benchmark. Increasing training data by 100× improves both, but unevenly: the agent is modeled well, while object fidelity remains far lower and improves slowly. We show that the agent gains need not come from data, and a careful visual conditioning design saturates fidelity with a fraction of it, which lets us measure object fidelity on its own and discover its saturation point. We then introduce a supervision scheme that shifts capacity from scene appearance toward object dynamics, improving object fidelity though a substantial gap remains. Finally, our conclusions transfer to downstream humanoid modeling. Overall, our results suggest that scaling ego-centric data brings agent modeling close to its limit while leaving its effects on the world far behind, and that closing this gap will depend on how models are trained, not only on how much data they see.
\end{abstract}

%% file: sections/1_intro.tex
\section{Introduction}

 World models~\citep{schmidhuber1990making} are being pursued for two distinct roles in embodied AI. As simulators, they promise to replace hand-authored environments for training and evaluating policies at scale, where relying on the physical world alone is prohibitively costly, slow, and unsafe~\citep{guo2026ctrl,gao2026dreamdojo}. As predictive models for planning, they promise to let an agent evaluate candidate actions at inference time by rolling out their consequences~\citep{ha2018world,bar2025navigation}. Both roles rest on the same capability: predicting how the objects an agent interacts with respond to its actions. This is precisely where conventional physics-based simulators fall short: contact, friction and deformation resist accurate simulation even when contact parameters are fit to real-world data~\citep{acosta2022validating,blanco2024benchmarking}. World models offer to learn these phenomena directly from data, if they in fact acquire them.

Recent efforts have scaled world models to large and diverse datasets of human–object interaction~\citep{punamiya2026egoverse}, in some cases comprising decades of interaction-dense footage~\citep{luo2026being,dyna2026}. Consistent gains with more data and larger models have encouraged a scaling-centric view: that dataset and model size are all that is needed to learn the dynamics of the world. Critics counter that pixel-level losses allocate model capacity by pixel mass and predictability rather than by causal relevance, so that the objects and contacts are precisely what the objective under-weights~\citep{lecun2022path}. If correct, this is a failure mode that scaling alone will not efficiently resolve. The argument has motivated alternative, latent-predictive formulations~\citep{assran2023self,bardes2024revisiting}, but the effect it posits has not been measured directly.

Testing that claim requires measuring interaction fidelity itself, across scale. Evidence for the scaling-centric view comes largely from downstream, task-level results: agreement between policy performance in learned and physics-based simulators~\citep{gao2026dreamdojo}, or offline action error when world models are used as policies~\citep{dyna2026}. Such evaluations compound errors across perception, agent motion, object dynamics, and policy behavior. We instead directly employ interaction-fidelity metrics~\citep{bagchi2026walk,pallotta2026egocontrol}, evaluating agent motion and its effect on the world separately as a function of training data, on a new out-of-distribution benchmark.

Specifically, we train Cosmos~3 video world models~\citep{nvidia2026cosmos3} on nested subsets, from 300 to 30,000 hours, of a large and diverse corpus of ego-centric human manipulation spanning 32k tasks and 116 environment types, collected by 14k participants (see Section~\ref{sec:trainset} for details). We repeat this sweep for three model variants and two action-conditioning designs, one of which supplies the agent's body configuration as a projected skeleton~\citep{wang2025precise,wu2026oscar}. All models are evaluated on an out-of-distribution benchmark with different capture hardware, sites, and participants, which we will release publicly.

Our experiments demonstrate that scaling ego-centric data improves both the agent and its action effects, but not equally. As shown in Figure~\ref{fig:agent_object_gap}, agent fidelity climbs steadily toward usable levels while object fidelity lags far behind, improving more slowly and from a lower base. Better conditioning substitutes for most of the agent gains: 1k hours of skeleton-conditioned training exceed the agent fidelity that standard conditioning reaches at 30k, and both converge to nearly the same limit (Figure~\ref{fig:data_scaling}, left). Saturating the agent this way isolates action effects and reveals that they converge far below the agent's fidelity and far below the level that is usable for simulation (Figure~\ref{fig:data_scaling}, right). 

Recently proposed interaction-focused objectives~\citep{zhang2026physisforcing} do not change this. We introduce object-centric adaptive noise scheduling
(Section~\ref{sec:object_ceiling}), which shifts supervision away from overall scene appearance and toward the dynamics of manipulated objects, and is the only intervention we test that raises the level object fidelity converges to. The improvement is nonetheless partial: object fidelity remains far below the agent's, and closing the remaining gap will require more than data. Finally, we show that these gains transfer to a downstream humanoid setting (Section~\ref{sec:humanoid}).

\paragraph{Contributions.} We present a controlled study of how interaction fidelity in ego-centric world models behaves as a function of scale. Concretely:
\begin{itemize}[leftmargin=*,itemsep=2pt,topsep=2pt]
    \item We train Cosmos~3 world models on up to 30{,}000 hours of human--object interaction and find that agent fidelity and the fidelity of the agent's effects on the world behave very differently under scale: the agent  saturates early and is reached far more cheaply through conditioning than through data, while its effects converge far below it (Section~\ref{sec:scale}).

    \item We propose object-centric adaptive noise scheduling, which shifts the training signal from scene appearance toward the dynamics of manipulated objects. It is the only intervention we test that raises the level object fidelity converges to, though a substantial gap remains (Section~\ref{sec:object_ceiling}).

    \item We show that these gains transfer to a downstream humanoid setting, where our design contributes more than ego-centric pre-training alone does, but modeling humanoid's interactions with the world remains a bottleneck (Section~\ref{sec:humanoid}).
\end{itemize}

%% file: sections/2_related.tex
\section{Related work}
\label{sec:related}
\paragraph{World models for robotics.}
The idea that an agent can act by consulting an internal model of its environment dates to \citet{craik1943nature}, and was realized in neural controllers that predict future observations and plan through them~\citep{schmidhuber1990making,sutton1991dyna} before being revived with modern architectures by \citet{ha2018world}. In robotics, such models served as forward predictors in MPC, with actions selected by sampling candidate sequences
and scoring their outcomes~\citep{finn2017deep,ebert2018visual}, or by optimizing policies within a learned latent space~\citep{hafner2019learning,hafner2023mastering}. Their scope was limited to
fixed setups with a single embodiment and few objects, by the small scale of available robotics datasets~\citep{dasari2019robonet} and by early generation
objectives~\citep{srivastava2015unsupervised,mathieu2016deep} that collapsed multi-modal futures toward a conditional mean and did not scale beyond low resolution and short horizons~\citep{villegas2019high}.

Both limitations have since been addressed. Diffusion
models~\citep{sohl2015deep,ho2020denoising} replaced regression with a sampling-based objective, and operating in a compressed latent space~\citep{rombach2022high} with transformer backbones~\citep{peebles2023scalable} made training on large datasets~\citep{chen2024panda} tractable, while coordinated collection efforts assembled robot interaction at a scale unavailable to earlier work~\citep{o2024open,khazatsky2024droid}. This resulted in a new wave of
world models for robotics: IRASim~\citep{zhu2025irasim} and Ctrl-World~\citep{guo2026ctrl} for tabletop manipulation, NWM~\citep{bar2025navigation} and EgoWM~\citep{bagchi2026walk} for
ego-centric navigation and manipulation. What still bounds them is the data: teleoperation remains expensive, and the largest datasets cover only a few hundred scenes~\citep{khazatsky2024droid}.

Ego-centric human video, in contrast, is abundant, diverse, requires no robot, and has driven a rapid scale-up of world-model pretraining.  DreamDojo~\citep{gao2026dreamdojo} trains on 44k hours before post-training on small amounts of target robot data, reporting strong action
controllability across several humanoid embodiments; Being-H0.7~\citep{luo2026being} scales to 200k hours in a world-action formulation; and EgoScale~\citep{zheng2026egoscale} uses diverse ego-centric human data to improve dexterous manipulation. Industrial systems report pretraining
on larger datasets still~\citep{dyna2026}, with the prevailing conclusion that more human video yields better world models.

\paragraph{Evaluating world models.} These conclusions rest on aggregate measures, such as agreement between policy rankings computed in the model and in the real world~\citep{zhu2025irasim,guo2026ctrl,gao2026dreamdojo} or action prediction accuracy~\citep{luo2026being,dyna2026}, which confirm practical utility but compound perception, dynamics, and control into a single number.  Model quality itself is otherwise reported with metrics inherited from video generation: frame-level SSIM~\citep{wang2004image}, PSNR~\citep{mannos1974effects}, LPIPS~\citep{zhang2018unreasonable} and DreamSim~\citep{fu2023dreamsim}, or distributional FVD~\citep{unterthiner2018towards}, which reward perceptual quality rather than accurate dynamics. None of these localize what improved.
Very recently, segmentation-based metrics that directly measure structural agreement between predicted and ground-truth videos have been proposed~\citep{bagchi2026walk,pallotta2026egocontrol}. We adopt this formulation, and report it separately for agents and objects being interacted with.

\paragraph{Learning interaction dynamics from pixels.}
A parallel line of work questions whether pixel prediction is the right objective for world modeling at all. \citet{lecun2022path} argues that predicting in observation space forces a model to allocate capacity to detail that is unpredictable and irrelevant to action, motivating joint-embedding architectures~\citep{assran2023self,bardes2024revisiting,assran2025v}. If this critique is correct, capacity should flow to what dominates the frame and is easiest to predict (e.g. the background) rather than to the small, frequently occluded objects where the interaction actually happens. In this work, we test this hypothesis directly at scale.

Several approaches to improving agent and interaction fidelity have been proposed. On the agent side, skeleton-based conditioning has been explored~\citep{wang2025precise,wu2026oscar}, motivated by cross-embodiment transferability, though its effect on scaling has not been examined. On the object side, earlier work proposed explicit object-centric representations~\citep{ferraro2025focus,feng2025fioc}, or auxiliary generation targets~\citep{yan2026open}, however, those sacrifice the scalability of generic diffusion architectures and do not generalize to ego-centric video. PhysisForcing~\citep{zhang2026physisforcing} adds trajectory and semantic alignment losses over physics-informative regions, but is evaluated on third-person robot manipulation with a fixed camera and largely unoccluded
interactions.  We find it yields no improvement at scale in our setting, and propose an approach that does not rely on those conditions, raising the level object fidelity converges to.

%% file: sections/4_protocol.tex
\section{Experiment Protocol}
\label{sec:protocol}
We train three variants of the Cosmos~3 video diffusion model~\citep{nvidia2026cosmos3}, differing in both design and size (Section~\ref{sec:preliminaries}), each with and without
skeleton conditioning, on nested data subsets spanning 300 to 30{,}000 hours (Section~\ref{sec:trainset}). All configurations are evaluated on an out-of-distribution benchmark drawn from a separate recording campaign, measuring how accurately the agent's body and the manipulated objects are rendered, with perceptual metrics as a control (Section~\ref{sec:eval}). Training details
are provided in appendix~\ref{sec:impl}.

\label{sec:method}
\begin{figure}[t]
 \centering
 \includegraphics[width=\linewidth]{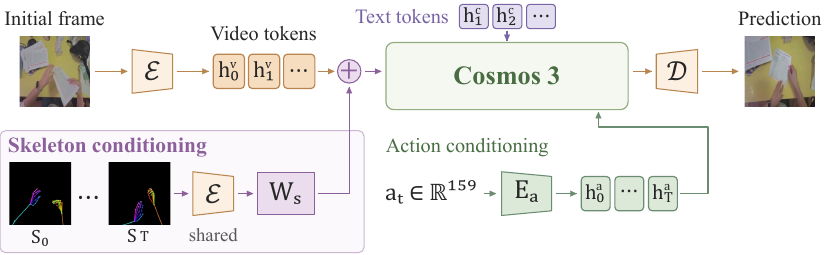}
 \caption{\textbf{Cosmos~3 with skeleton conditioning.} The projected skeleton sequence $S_{0:T}$, shown in bottom left, is encoded by the same frozen VAE $\mathcal{E}$ as the video, and its latent patches are projected through a zero-initialized $W_s$ and added to the corresponding video embeddings. The skeleton is derived from the camera intrinsics to better align agent's actions with the visual token grid.}
 \label{fig:method_overview}
 \vspace{-10px}
\end{figure}

\subsection{World Model Design}
\label{sec:preliminaries}
We build on the pretrained Cosmos~3 video diffusion model~\citep{nvidia2026cosmos3} (Figure~\ref{fig:method_overview}). Given an initial RGB observation $x_0$, a text description $c$, and an agent-motion sequence $a_{0:T}$, it predicts future frames $x_{1:T}$. A frozen video
autoencoder $\mathcal{E}$ maps the video to spatiotemporal latents, which a dual-tower backbone processes together with text and action tokens: text attends causally, while video and action tokens attend jointly to one another and to the text context. We study three model variants. Edge is a 4B model whose 2B denoising transformer is custom-designed and trained from scratch, while Nano and Super are 16B and 64B models, respectively. Their 8B and 32B transformers are adapted from pretrained Qwen3-VL~\citep{bai2025qwen3} variants of the same size. Edge therefore differs from the other two in pretraining as well as design, and we treat comparisons involving it as comparisons between model variants rather than as a capacity trend. Nano and Super share an initialization and differ only in scale, so their comparison does isolate capacity. Due to the high cost of training Super we report it with skeleton conditioning only

We extend the model's action space with full hand articulation and shoulder/elbow positions:
\[
s_t =
\big[
\underbrace{c_t}_{9},
\underbrace{w_t^L,w_t^R}_{18},
\underbrace{h_t^L,h_t^R}_{120},
\underbrace{b_t}_{12}
\big]\in\mathbb{R}^{159},
\]
where $c_t$ and $w_t^{L/R}$ are camera and wrist poses, each with 3D translation and 6D rotation; $h_t^{L/R}$ each hold 20 hand-joint positions in the corresponding wrist's frame and $b_t$ holds both shoulders and elbows. Actions $a_t$ are obtained by normalizing $s_0$ and the element-wise differences $s_t-s_{t-1}$; a learned linear encoder maps each to one action token, which remains clean during denoising.

We train the diffusion branch and its conditioning modules with rectified flow
matching~\citep{liu2023flow}, keeping the VLM and VAE fixed. Let $z=\mathcal{E}(x_{0:T})$ be the
clean video latents, $\epsilon\sim\mathcal{N}(0,I)$, and $M^{\mathrm{pred}}$ a binary mask
selecting future latents. At noise level $\sigma\in[0,1]$,
\begin{equation}
\begin{aligned}
z_\sigma &= z+\sigma M^{\mathrm{pred}}\odot(\epsilon-z),\\
\mathcal{L}_{\mathrm{FM}} &= \mathbb{E}\!\left[
\left\|M^{\mathrm{pred}}\odot\left(v_\theta(z_\sigma,\sigma;c,a_{0:T})-(\epsilon-z)\right)\right\|_2^2
\right].
\end{aligned}
\label{eq:flow_matching}
\end{equation}
The initial observation remains clean. At inference, the observed latents are fixed, and we decode the resulting video latents with the VAE decoder $\mathcal{D}$.

\paragraph{Skeleton conditioning.}
Action tokens describe agent motion in a coordinate space of poses and joint angles, leaving the model to learn how it projects onto the pixels. We remove this inference by additionally
conditioning on a spatially aligned skeleton sequence (Figure~\ref{fig:method_overview}, bottom left), placing the agent's geometry on the video-token grid. For each frame we project
the two 21-joint hands and their shoulder--elbow--wrist chains onto the image plane: a joint at
camera-frame position $q_{t,j}$ has pixel coordinates $u_{t,j}=(K_tq_{t,j})_{1:2}/(K_tq_{t,j})_3$
and depth $d_{t,j}=(q_{t,j})_3$, with camera intrinsics $K_t$.

We rasterize the joints and connecting bones onto a black RGB canvas. Projection discards depth, so we encode it in colour: hue identifies which finger or limb a primitive belongs to, and value darkens with distance, so the skeleton carries 3D orientation as well as image-plane position. A primitive at depth $d_p$ (in meters, the mean endpoint depth for bones) is drawn in
\begin{equation}
\begin{aligned}
\eta(d_p)&=\operatorname{clip}\!\left(\frac{d_p-0.15}{0.70-0.15},0,1\right),\\
C_p&=\operatorname{HSVtoRGB}\!\left(
\frac{\phi_p}{360},\alpha_p,1-\beta_p\eta(d_p)
\right),
\end{aligned}
\label{eq:skeleton_color}
\end{equation}
where $\eta(d_p)$ maps depth onto a normalized shading factor over the working range, $\phi_p$ is the hue assigned to the primitive's finger or limb, and $(\alpha_p,\beta_p)$ control saturation and the strength of depth shading; exact values are given in Appendix~\ref{sec:impl}. 

To condition the diffusion model, we encode the rasterized sequence
$S_{0:T}$ with the same frozen VAE used for video and partition the
latents into patches aligned with the video grid.
We denote each vectorized skeleton latent patch by
$z_i^{\mathrm{skel}}$ and its projected token by
$h_i^{\mathrm{s}} = W_s z_i^{\mathrm{skel}}$, where $W_s$ is a
bias-free linear projection. We add this skeleton token to the corresponding video token as $\widetilde{h}_i = h_i^{\mathrm{v}} + h_i^{\mathrm{s}}$. The skeleton is supplied at both training and inference time.

\subsection{Training}
\label{sec:trainset}
\begin{figure}[t]
  \centering
  \includegraphics[width=\linewidth]{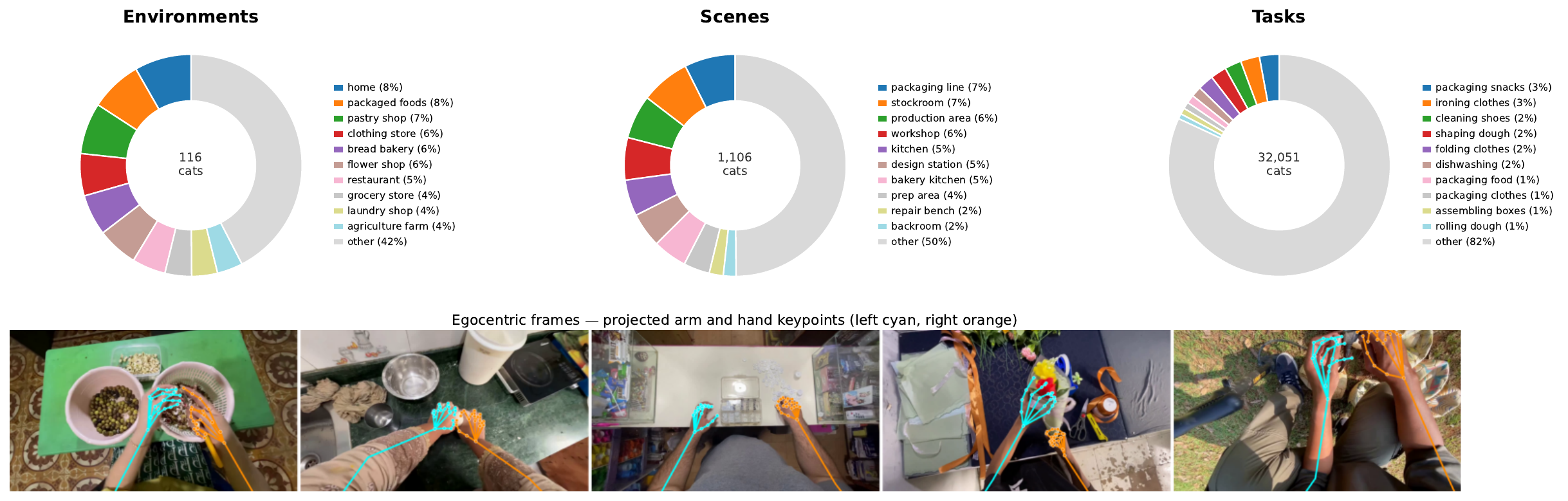}
  \caption{\textbf{Dataset overview.} \emph{Top:} distribution of clips over
  environments, scenes, and tasks (ten most frequent categories each; the
  gray wedge aggregates the long tail). \emph{Bottom:} representative
  first-person frames from four environments, overlaid with the dataset's
  per-frame keypoints.}
  \label{fig:dataset-overview}
  \vspace{-10px}
\end{figure}

We train all models on a large-scale ego-centric dataset of everyday human manipulation, collected by Mecka\footnote{\url{https://mecka.ai}}: 1.15M first-person clips totalling 30k hours ($3.24{\times}10^{9}$ frames at $30$\,fps, mean clip length $94$\,s), each recording a manipulation activity from a head-mounted camera. Every frame carries calibrated intrinsics and a $6$-DoF camera pose, dense 3D hand keypoints (21 per hand)
and wrist poses, upper-body keypoints, and a free-form activity description. Crucially for studying generalization, the corpus is highly diverse: 116 environment categories (homes, bakeries, flower shops, farms, repair shops), 1{,}106 scene categories and 32{,}051 tasks from 14{,}019 participants, with no environment type exceeding $8.3\%$ of the data. Figure~\ref{fig:dataset-overview} summarizes the environment, scene and task distributions as well as visualizes a few samples from the dataset. 

We train on a \emph{balanced, nested data ladder} at $300$, $1$k, $3$k, $10$k and $30$k hours. A fixed episode order balances joint environment--scene groups according to their video-duration shares in the original training pool. Each budget is a checkpoint along one progressive training run, so larger budgets extend the same data prefix and increase both data exposure and optimization compute. Appendix~\ref{app:data_scaling_ladder} details the ladder and defines the exposure in hours.

\subsection{Evaluation}
\label{sec:eval}
\paragraph{Dataset.}
Because we ask whether \emph{generic} models of object dynamics can be learned from ego-centric human video, we evaluate out of distribution. Our evaluation set comes from a separate recording campaign: none of its sessions appear in training, and since platform, operators and sites all differ, no location or participant is shared. $37\%$ of its environment categories, accounting for a quarter of its clips, never appear in training (Appendix~\ref{sec:app_eval}). From this evaluation set we hand-curated $150$ clips under three criteria: (i)~\emph{non-trivial agent--object interaction} --- each clip contains active manipulation rather than idle or transit motion;
(ii)~\emph{segmentable objects} --- the manipulated objects are large enough in
the image to be segmented accurately, which mask-based metrics rely on; and
(iii)~\emph{accurate body-joint annotations} --- clips with accurate upper-body
and hand keypoints only.

Our key metrics are computed over the regions occupied by the hands and the manipulated objects rather than the full frame. We annotate every clip with SAM\,2~\citep{sam2}, prompting a few points on the first frame for the hands and each manipulated object and propagating them with the video tracker, then manually verify and correct the resulting tracks. The evaluation set
comprises $150$ clips of $32$ frames at $10$\,fps with $11{,}328$ instance masks; models are evaluated on $17$-frame windows, one observed frame and $16$ predicted. We will release the entire evaluation set.

\paragraph{Metrics.}
We measure structural consistency and perceptual quality. SCS~\citep{bagchi2026walk} segments and tracks key scene elements through the predicted and ground-truth videos and averages mask IoU, capturing whether the provided actions are accurately reflected in the generated world. We
report it separately for the hands (agent) and the manipulated objects.
LPIPS~\citep{zhang2018unreasonable} serves as a perceptual control, so that differences in structural fidelity can be checked against differences in image quality.

Because mask tracking is itself imperfect, even a perfect prediction would not score an SCS of $1.0$. We estimate achievable maximum by re-initializing the tracker from different frames of the same video, with prompts sampled from the ground-truth masks, and measuring agreement between the resulting tracks: 0.93 for the agent and 0.90 for objects on ground-truth video, with comparable values on generated video, so the bound applies to model outputs too. Both lie well above the scores our models attain, so the differences we report reflect the models rather than the limits of the metric.

%% file: sections/5_experiment.tex
\section{How Far Does Ego-centric Data Take Us?}
\label{sec:scale}
\begin{figure}[!t]
\centering
\includegraphics[width=\linewidth]{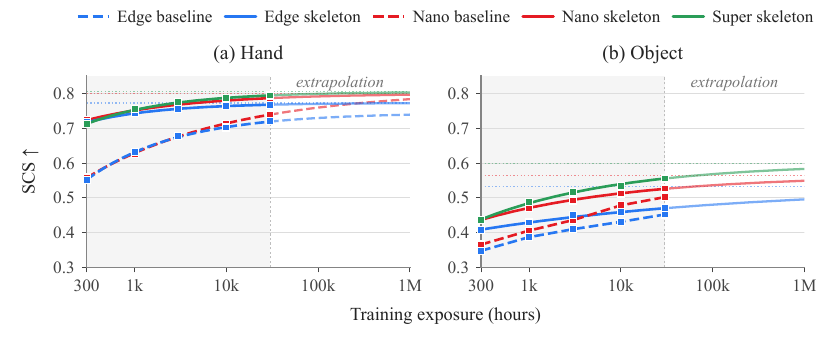}
\caption{\textbf{Scaling reaches the agent's limit, not the object's.} SCS for the agent (left) and manipulated objects (right) versus training data, with and without skeleton conditioning. Skeleton conditioning reaches the agent's limit with far less data, while object fidelity converges far below.}
\label{fig:data_scaling}
\vspace{-5px}
\end{figure}

\paragraph{Data scaling saturates agent fidelity.}
We first ask what data volume alone provides, measuring directly how accurately the model renders the agent's body and the objects it manipulates. Perceptual metrics are reported in Appendix~\ref{app:perceptual_scaling} as a control, confirming that the differences we measure are not confounded with image quality. The grey portion of the plots in Figure~\ref{fig:data_scaling} shows agent and object fidelity as a function of pretraining data for Edge and Nano Cosmos~3 variants (dashed lines). Both improve across the full range, although the gap between modeling the agent and modeling its effect on the world grows with data scaling. Nano achieves both higher absolute fidelity and faster improvement than Edge, most visibly on objects ($+0.14$ versus $+0.10$). Read at face value, this supports the scaling-centric view: fidelity rises with both data and model, suggesting continued scaling might eventually close the gap.

The two axes differ, however, in what the curves reveal about their limits. Agent fidelity is visibly saturating: it gains 0.12 SCS between 300 and 3k hours but only 0.06 between 3k and 30k, and fitting a saturating curve of the form $S_\infty - cD^{-\alpha}$, standard in scaling-law analyses~\citep{kaplan2020scaling,hoffmann2022training}, places its asymptote at 0.811. Object fidelity shows no comparable deceleration, and a curve without curvature does not constrain its asymptote, so the object limit remains undetermined within our range, which is why we plot these curves only over the observed range. Rather than extending data collection by another order of magnitude, we next use skeleton conditioning as an instrument for locating that limit.

\paragraph{Conditioning fixes the agent, revealing object scaling.}
\label{sec:scale_skeleton}
Skeleton conditioning supplies the agent's projected body configuration in the camera frame, so the model no longer has to infer it from action tokens. The effect on the agent axis is large (solid curves in Figure~\ref{fig:data_scaling}, left): with 300 hours, the skeleton-conditioned Nano variant reaches the agent fidelity the baseline attains only at roughly 15k hours - a $\sim$50$\times$ reduction in data - and gains just 0.06 SCS over the remaining two orders of magnitude. Crucially, conditioning changes \emph{when} the agent's limit is reached, not \emph{what} it is: the two designs converge to about 0.01 SCS of one another (0.811 and 0.800). Additional ego-centric video therefore cannot meaningfully improve this axis, and saturating it early gives us an instrument for locating the object limit within our budget.

Objects benefit from conditioning as well (Figure~\ref{fig:data_scaling}, right), the Nano variant reaching at 300 hours the baseline's fidelity at 3k, though this advantage narrows across the ladder. This could indicate a shortcut: by supplying body configuration directly, conditioning might spare the model visual inference that could support interaction modeling. We test this in Appendix~\ref{app:drop} by randomly dropping the skeleton during training, which retains that inference, and find no evidence of a shortcut. Conditioning therefore isolates the quantity we care about: with the agent held near its ceiling across the ladder, the object curve is largely unconfounded by improvements in agent fidelity.

Measured this way, object fidelity improves by only 0.09 SCS across two orders of magnitude, and unlike the baseline it decelerates, gaining 0.056 between 300 and 3k hours but 0.035 between 3k and 30k. The curve is now in its saturating regime and its asymptote is identifiable: 0.565, of which the model has already realized 93\% at 30k hours. Normalized by the maximum the metric can register for each category, the agent converges to 86\% of its measurable range and objects to 63\%. The limitation is therefore not that object dynamics improve slowly with data, but that the level they improve toward lies far below the agent's.  

Scaling the model size by a factor of 4 (Super variant shown in green) has virtually no effect on the agent fidelity, but raises the object ceiling by 0.034 SCS points. On one hand, this supports the scaling-centric view, suggesting that growing the model size together with the data could eventually bring interaction modeling fidelity to a usable level. Taken quantitatively, on the other hand, it does not. At 0.034 per quadrupling, closing the remaining 0.205 to the agent ceiling would require roughly 6 further quadruplings, some 4 orders of magnitude more parameters, with training compute rising in proportion. The estimate is a lower bound, since it assumes the per-quadrupling gain does not diminish. Neither axis we can scale therefore offers a practical route to closing this gap, so we turn to what the model is trained to predict.

\section{Can the object ceiling be moved?}
\label{sec:object_ceiling}
The previous section established a low limit on object fidelity that additional data is unlikely to overcome. We now ask whether changing the supervision can, using the Cosmos~3 Nano variant.

\begin{figure}[t]
\centering
\begin{minipage}[t]{0.51\linewidth}
\vspace{0pt}
\begin{minipage}[t][1.78in][c]{\linewidth}
\centering
\begingroup
\fontsize{8}{9.4}\selectfont
\setlength{\tabcolsep}{2pt}
\renewcommand{\arraystretch}{1.22}
\newcommand{\tableonevalue}[2]{%
  #1\,{\fontsize{7}{8.2}\selectfont
  \textcolor[gray]{0.45}{$\pm$\,#2}}%
}
\resizebox{\linewidth}{!}{%
\begin{tabular}{@{}lrrr@{}}
\toprule
Variant
& \multicolumn{1}{c}{Agent $\uparrow$}
& \multicolumn{1}{c}{Object $\uparrow$}
& \multicolumn{1}{c@{}}{LPIPS $\downarrow$} \\
\midrule
Skeleton
& \tableonevalue{0.779}{0.002}
& \tableonevalue{0.513}{0.004}
& \tableonevalue{0.235}{0.002} \\
\midrule
\multicolumn{4}{@{}l}{\textit{Prior auxiliary objectives}} \\
\quad PhysisForcing
& \tableonevalue{0.780}{0.002}
& \tableonevalue{0.515}{0.005}
& \tableonevalue{0.234}{0.002} \\
\midrule
\multicolumn{4}{@{}l}{\textit{Supervision allocation}} \\
\quad Reweighting
& \tableonevalue{0.781}{0.002}
& \tableonevalue{0.516}{0.004}
& \tableonevalue{0.235}{0.001} \\
\quad Dynamic noise
& \tableonevalue{0.786}{0.001}
& \tableonevalue{0.526}{0.004}
& \tableonevalue{0.234}{0.002} \\
\mbox{\quad Dynamic noise (random)}
& \tableonevalue{0.781}{0.002}
& \tableonevalue{0.517}{0.005}
& \tableonevalue{0.233}{0.002} \\
\rowcolor[RGB]{240,246,243}
\quad \textbf{Ours}
& \tableonevalue{\textbf{0.788}}{0.002}
& \tableonevalue{\textbf{0.532}}{0.003}
& \tableonevalue{0.235}{0.002} \\
\bottomrule
\end{tabular}%
}
\endgroup
\end{minipage}
\captionof{table}{\textbf{Regional supervision ablation at 10k hours.}
Combining dynamic noise scheduling and loss reweighting achieves the highest object fidelity, without sacrificing perceptual quality. Mean $\pm$ SD across five seeds.}
\label{tab:ablation}
\end{minipage}\hfill
\begin{minipage}[t]{0.47\linewidth}
\vspace{0pt}
\begin{minipage}[t][1.78in][c]{\linewidth}

\includegraphics[width=\linewidth]{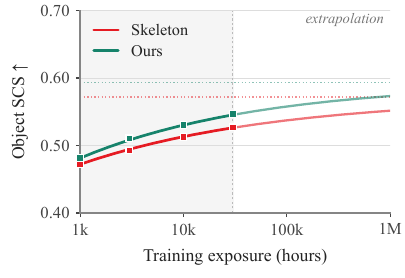}
\end{minipage}
\captionof{figure}{\textbf{A higher object ceiling.} Object fidelity across the data ladder, with and without dynamic-region supervision. Ours exceeds the skeleton baseline at every budget, and converges to a higher limit.}
\label{fig:method_scaling}
\end{minipage}
\vspace{-10bp}
\end{figure}

\paragraph{Existing interaction-focused supervision does not transfer.}
PhysisForcing~\citep{zhang2026physisforcing} localizes `physics-informative' regions using 2D point tracking and monocular depth, then within them trains intermediate diffusion features to support point tracking against CoTracker3~\citep{karaev2025cotracker3} references and aligns
their pairwise token-similarity matrix with a frozen V-JEPA~2
encoder~\citep{assran2025v}. Re-implementing both objectives, we find they do not improve over the skeleton-conditioned baseline at 10k hours (Table~\ref{tab:ablation}). Both the localization heuristic and the tracking supervision rely on a static camera and largely unoccluded objects, and in ego-centric manipulation neither holds: the camera moves with the
agent, and the object is occluded by the very hand acting on it at the moment of contact. Next, we design an approach that depends on neither assumption.

\paragraph{Localizing dynamic regions.}
As shown in Figure~\ref{fig:object_supervision} (a), we first initialize a regular grid of queries in the first frame and track them through the video with D4RT~\citep{zhang2026efficiently}. Combining each point's 3D displacement and tracking reliability with a soft spatial prior around the projected hands, and expanding the resulting scores over neighbouring latent cells, yields a dynamic region map $M^{\mathrm{dyn}}\in[0,1]^{H\times W}$
(see Appendix~\ref{app:dynamic_region_details} for details).

\paragraph{Shifting the supervision.}
Within these regions we raise the noise level, so the denoising target can no longer be satisfied by propagating local appearance and must be predicted from context and dynamics. With $\Omega=\{i:M_i^{\mathrm{pred}}=1\}$ the predicted latent locations, $z_\sigma$ from Equation~\ref{eq:flow_matching} and $v=\epsilon-z$, for $i\in\Omega$ and $0<\sigma<1$:
\begin{equation}
\delta_i = \lambda M_i^{\mathrm{dyn}}\min(\sigma,1-\sigma),
\qquad
\widetilde z_{\sigma,i} = z_{\sigma,i}+\delta_i v_i,
\qquad
\widetilde v_i=\left(1+\tfrac{\delta_i}{\sigma}\right)v_i,
\label{eq:paired_regional_corruption}
\end{equation}
where $\lambda\in[0,1]$ scales the additional noise in the selected regions. This raises the effective noise level from $\sigma$ to $\sigma+\delta_i$ while retaining $\sigma$ as the model's noise-level input, and the paired target satisfies $\widetilde z_{\sigma,i}-\sigma\widetilde v_i=z_i$, so supervision still points toward the same clean latent. For comparison we also reweight the loss within the same
regions (described in Appendix~\ref{app:dynamic_region_details}).

\paragraph{Results.}
We first report an ablation of different components of our objective at 10k hours in Table~\ref{tab:ablation}. Reweighting alone does not meaningfully improve fidelity, whereas raising the noise level does,
and is the only intervention we test that improves object fidelity. Replacing $M^{\mathrm{dyn}}$ with randomly placed regions of the same extent removes nearly all of the gain, confirming that the benefit comes from where supervision is shifted rather than from the perturbation itself. Reweighting adds a small further improvement at no extra cost, so we include it in our final configuration. LPIPS is unchanged throughout, confirming that our approach does not sacrifice perceptual quality.

Figure~\ref{fig:method_scaling} reports object interaction fidelity across the data ladder. Our method exceeds the skeleton baseline at every budget, reaching 0.546 against 0.527 at 30k hours (improvement of 3.7\%), and fitting the same saturating form places its asymptote above the baseline's. The gain is, however, roughly constant across budgets: the ceiling is raised, not the rate at which it is approached, and it remains far below the agent's. Improving interaction fidelity to a usable level will require more substantial changes to how world models are trained.

\section{Transfer to humanoids}
\label{sec:humanoid}

\paragraph{Setup.}
We fine-tune Cosmos~3 Nano on ground-truth demonstrations from 11 tasks from the simulated bimanual benchmark of~\citet{yang2025egovla}, conditioning on a single start frame and the 138-dimensional humanoid state trajectory and predicting the following 16 frames (17 total at 10\,Hz). We compare three variants sharing the same recipe and evaluation clips: \textsc{Cosmos~3}, from the released weights, which have never seen our ego-centric data; \textsc{+Human}, from the matched checkpoint pre-trained on 30{,}000\,h without skeleton conditioning or our supervision scheme;
and \textsc{Ours}, from the checkpoint pre-trained with both, which remain active during fine-tuning. The first pair isolates ego-centric human pre-training, the second our design. All evaluations are performed on EgoVLA policy rollouts rather than demonstrations. Further details provided in Appendix~\ref{app:humanoid}.

\paragraph{Metrics.}
We report SCS for hands and manipulated objects separately, using manually verified masks. To assess applicability to downstream robotics evaluation, we additionally measure how well each model's rendered rollouts support the same success judgment as the simulator's: within each rollout we locate the 17-frame window in which the outcome is decided, replay its actions through each variant, and apply the benchmark's success criterion to both renderings, reporting
how often the two verdicts agree. Restricting to the critical window isolates world-model fidelity from drift over long rollouts. Appendix~\ref{app:humanoid} gives the full protocol, and per-task results.

\paragraph{Results}
Ego-centric human pre-training transfers, improving every measure and most clearly hand SCS ($+0.07$, Table~\ref{tab:humanoid-ablation}). Our conditioning and supervision design contributes more than twice as much on top of it. The pattern matches our human-video findings in one respect and departs from it in another. As there, gains are larger for the agent than for the objects it manipulates ($+0.17$ versus $+0.06$). Unlike there, object fidelity is high in
absolute terms for every variant ($0.70$–$0.79$), which we attribute to the simulated environment: its objects are rigid and relatively large, so the conditions that make object dynamics hard in ego-centric human video are largely absent. Conclusions about interaction fidelity thus depend heavily on the difficulty of the evaluation setting.

Success agreement follows the same ordering with smaller margins, and is less diagnostic: a single number records that agreement differed without indicating where. SCS localizes the residual error to the objects. Over the prediction horizon, hand SCS for \textsc{Ours} is nearly flat, falling from 0.89 at frame~2 to 0.85 at frame~16, while object SCS falls more than twice as far, from 0.82 to 0.71. The model knows where the hands will be, but fails to correctly predict the consequences of agent's actions in the world.  We illustrate this via qualitative analysis in Appendix~\ref{app:qual_humanoid}.

\begin{table}[t]
\centering
\caption{\textbf{Transfer to humanoid manipulation.} SCS on the humanoid benchmark, and agreement with the simulator's rendering on policy-rollout success. Contribution columns give the difference between adjacent variants. Higher is better.}
\label{tab:humanoid-ablation}
\setlength{\tabcolsep}{6pt}
\renewcommand{\arraystretch}{1.1}
\begin{tabular}{@{}lccc@{\hskip 16pt}rr@{}}
\toprule
 & \multicolumn{3}{c}{Variant} & \multicolumn{2}{c@{}}{Contribution} \\
\cmidrule(lr){2-4}\cmidrule(l){5-6}
Metric & \textsc{Cosmos~3} & \textsc{+Human} & \textsc{Ours}
       & Human pre-train & Ours \\
\midrule
Hand SCS          & 0.63 & 0.70 & \textbf{0.88} & $+0.07$ & $+0.17$ \\
Object SCS        & 0.70 & 0.73 & \textbf{0.79} & $+0.03$ & $+0.06$ \\
\addlinespace[2pt]
Success agreement & 0.67 & 0.71 & \textbf{0.81} & $+0.04$ & $+0.10$ \\
\bottomrule
\end{tabular}
\vspace{-12px}
\end{table}

%% file: sections/6_conclusion.tex
\section{Discussion and Limitations}
Our results are not an argument against collecting ego-centric human video: more data improved both axes at every scale we measured, and pre-training transferred to humanoids. They are an argument against expecting collection alone to close the interaction gap. Agent fidelity approaches its limit well before 30k hours, while the fidelity of the agent's effects on the world converges far below it, even with our supervision scheme.

Two factors plausibly contribute (see Appendix~\ref{app:qual_human} for qualitative analysis). The first is irreducible uncertainty: how an object responds to contact depends on mass, friction and stiffness, none of which is perfectly recoverable from a single frame. The second is that the quantities that matter most are not observable in pixels at all. An object's geometry is only partially visible, and the parts most relevant to an interaction (those under the hand, at the point of contact) are exactly the parts the hand occludes. Contacts and forces are never visible. These are precisely what a physics simulator represents exactly
and can supply as supervision, while remaining poor at the complex dynamics that motivated learned world models in the first place. Each paradigm is strongest where the other is weakest, and recent work distilling simulators into differentiable generative 3D particle models~\citep{wang2026physctrl,duisterhof2026pointzero} suggests a path to combining them.

\paragraph{Limitations.} Our data ladder spans two orders of magnitude and couples data volume with optimization budget. It therefore establishes where fidelity is heading under joint scaling, not the effect of either factor alone.
However, because the smaller budgets receive proportionally fewer optimization steps, they are further from convergence than the larger ones, and training each to convergence would raise the left of the curve more than the right and therefore lower the fitted asymptote. The ceilings we report are in this sense optimistic. They also hold only insofar as the fitted regime continues beyond 30k hours. Despite their diversity, both our training and evaluation sets come from a single collection protocol, and the evaluation set is limited in scale, which constrains how far our conclusions generalize to ego-centric video as a whole. Finally, our experiments study generative world models. Whether latent-predictive models learn to capture interaction dynamics more readily remains open, and cannot be settled with our fidelity metrics as those models do not render pixels.

%% file: sections/a_appendix.tex
\section{Training Set Details}
\label{app:train_set}

\subsection{Overview and modalities}
Our training set is a large-scale collection of egocentric video of humans performing
everyday manipulation activities, recorded from a head-mounted camera. Each clip
is a single continuous recording (mean length $94.3$\,s, median $94.5$\,s). We
use the \emph{full} corpus of $1{,}146{,}100$ clips ($30{,}012$ hours,
$3.24{\times}10^{9}$ RGB frames at $30$\,fps). Every clip is annotated per frame
with the following aligned modalities:
\begin{itemize}
  \item \textbf{RGB} egocentric video (undistorted pinhole projection).
  \item \textbf{Camera calibration}: per-frame intrinsics and a $6$-DoF camera
        pose (camera-to-world, metric scale) recovered by SLAM.
  \item \textbf{Hands}: $3$D keypoints for both hands ($21$ landmarks each) and
        $6$-DoF wrist poses, expressed in the camera frame.
  \item \textbf{Body}: upper-body keypoints in a SMPL-H--compatible layout.
  \item \textbf{Language}: a task label and a free-form natural-language
        description of the activity.
\end{itemize}

\subsection{Scale and diversity}
\begin{figure}[t]
  \centering
  \includegraphics[width=\linewidth]{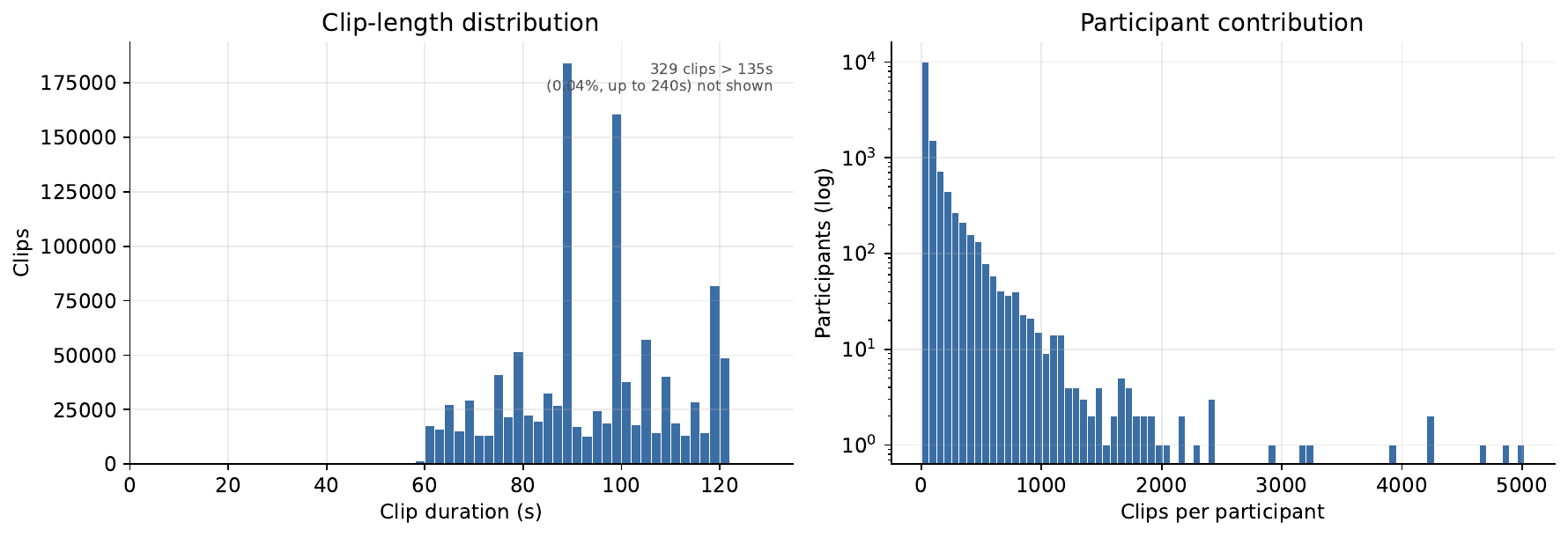}
  \caption{\textbf{Left:} clip-length distribution (mean $94.3$\,s, median
  $94.5$\,s; $99.9\%$ of clips are $\le 123$\,s). The axis is truncated at
  $135$\,s; a sparse tail ($0.04\%$ of clips) extends to a $240$\,s cap and is
  omitted. \textbf{Right:} histogram of clips per participant (log $y$-axis),
  showing a heavy tail across $14{,}019$ contributors.}
  \label{fig:duration-participants}
\end{figure}

Table~\ref{tab:dataset-stats} reports headline statistics. The corpus is
diverse along every axis we measured: $116$ environments, $1{,}106$ scenes,
$32{,}051$ tasks, and $14{,}019$ participants. Importantly, coverage is not
concentrated in a few dominant categories. The most common environment category (\emph{home}) accounts for only $8.3\%$ of clips; reaching $90\%$ coverage requires $34$ of the $116$ environment categories, $139$ scene categories, and $3{,}745$ task categories. Participation is likewise long-tailed: the median
participant contributes $19$ clips while the most prolific contributes $5{,}019$,
and the top $1\%$ of participants account for $17\%$ of all clips
(Fig.~\ref{fig:duration-participants}, right).
\begin{table}[t]
  \centering
  \caption{Training set summary statistics (full corpus).}
  \label{tab:dataset-stats}
  \begin{tabular}{lr}
    \toprule
    Statistic & Value \\
    \midrule
    Clips (recordings)        & $1{,}146{,}100$ \\
    Total duration            & $30{,}012$ hours \\
    Total frames ($30$\,fps)  & $3{,}241{,}303{,}112$ \\
    Mean / median clip length & $94.3$\,s / $94.5$\,s \\
    Clip length range         & $2.6$--$240.0$\,s \\
    \midrule
    Unique environments       & $116$ \\
    Unique scenes (merged)    & $1{,}106$ \\
    Unique tasks (merged)     & $32{,}051$ \\
    Unique participants       & $14{,}019$ \\
    Clips / participant (med., max) & $19$, \; $5{,}019$ \\
    \bottomrule
  \end{tabular}
\end{table}

\subsection{Category distributions}
\begin{figure}[t]
  \centering
  \includegraphics[width=\linewidth]{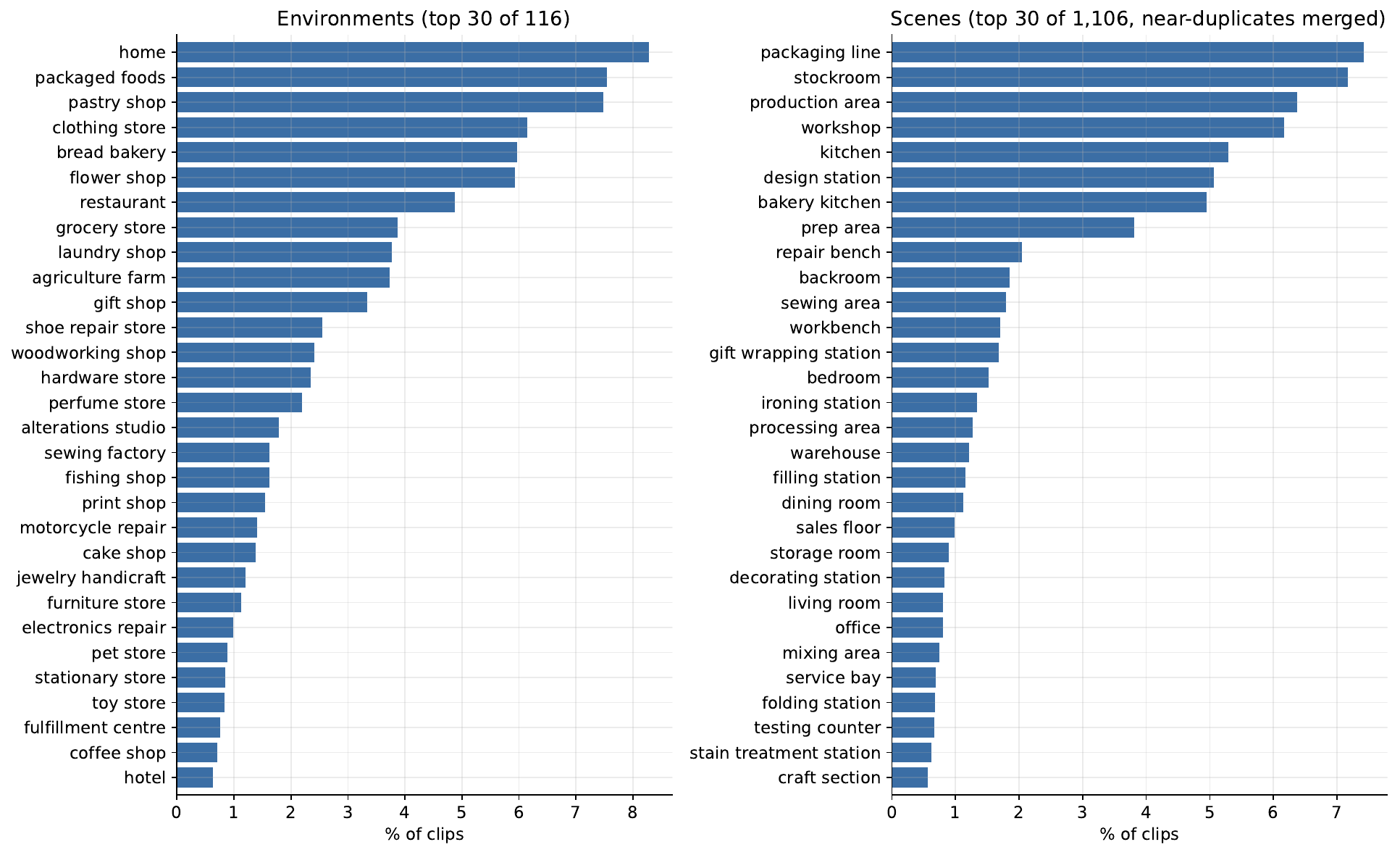}
  \caption{Most frequent \textbf{environments} (left, top $30$ of $116$) and
  \textbf{scenes} (right, top $30$ of $1{,}106$, near-duplicates merged), as a percentage of all clips.}
  \label{fig:env-scene}
\end{figure}

\begin{figure}[t]
  \centering
  \includegraphics[width=0.99\linewidth]{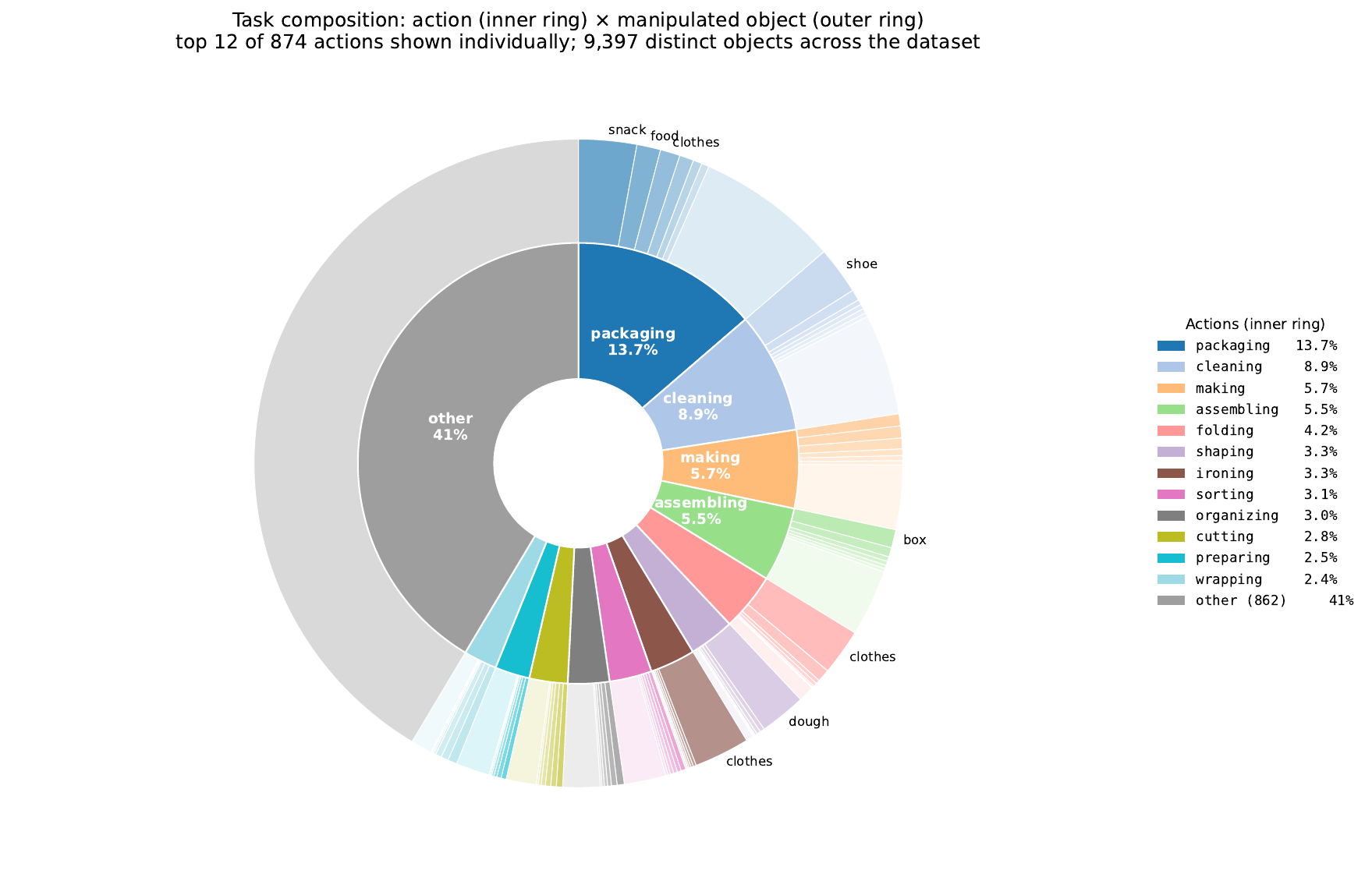}
  \caption{\textbf{Task composition.} Each task label decomposes into an action
  (inner ring) applied to a manipulated object (outer ring); together the corpus
  spans $874$ distinct actions and $9{,}397$ distinct objects ($32{,}051$
  action--object tasks). For legibility the inner ring shows the $12$ most
  frequent actions with wedge angles proportional to their share of all clips;
  the remaining $862$ actions ($41\%$ of clips) are grouped into the grey
  ``other'' wedge. The outer ring shows the top objects within each action, with
  that action's remaining objects aggregated into a lighter ``other'' sub-wedge.
  While a few actions (packaging, cleaning, making) are common, each fans out
  over a broad range of objects.}
  \label{fig:tasks}
\end{figure}

Figure~\ref{fig:env-scene} shows the $30$ most frequent environments and scenes,
and Figure~\ref{fig:tasks} decomposes tasks into actions and manipulated objects. The environment mix spans domestic settings (homes, kitchens,
bedrooms), food production (bakeries, pastry and bread shops, restaurants),
retail (clothing, grocery, gift, and hardware stores), and craft/repair
workshops (woodworking, sewing, shoe and motorcycle repair), among others. Tasks
are dominated by fine bimanual manipulation such as ironing and folding clothes,
shaping and rolling dough, cleaning, and packaging, consistent with the
free-form descriptions in which \emph{small}, \emph{plastic}, \emph{placing},
\emph{folding}, and \emph{cutting} are among the most frequent content words.

\subsection{Balanced data scaling ladder}
\label{app:data_scaling_ladder}

\paragraph{Balanced ordering.}
We construct an episode order by grouping episodes that share the same environment and scene labels. Episodes within each group are ordered deterministically using a fixed seed, and the groups are interleaved according to their shares of video duration in the original training pool. Specifically, let $T_g$ be the total duration of group $g$ in this reference pool, $a_g$ the duration already selected from that group, and $d_g$ the duration of its next episode. Among groups with remaining episodes, we select
\begin{equation}
g^*=\arg\min_g\frac{a_g+d_g}{T_g}.
\end{equation}
This keeps each group's share of the selected video duration approximately equal to its share in the reference pool as more episodes are selected. Overall, the balance is approximate because episodes have different durations and the available data in each group are finite.

\paragraph{Nested budgets.}
We use the same episode order for all model configurations and construct the training examples. The budgets $300$, $1$k, $3$k, $10$k and $30$k are reached by processing increasing numbers of samples in this fixed order, so each larger budget includes all samples used at smaller budgets. For each configuration, the model uses each sample once, and data exposure and optimizer updates therefore increase together.

\paragraph{Exposure in hours.}
Let $N=33{,}512{,}693$ be the total number of training examples in the fixed manifest and $n$ the number consumed. We define
\begin{equation}
H(n)=30{,}000\,\frac{n}{N}\;\text{hours},
\qquad n=Bs,\quad B=512,
\label{eq:clip_equivalent_hours}
\end{equation}
where $s$ is the optimizer step and $B$ is the global batch size. Thus, hours measure the fraction of the corpus consumed, normalized to its nominal $30{,}000$-hour size; they do not sum the durations of the sampled $17$-frame windows. We align each budget to a whole optimizer step and report rounded hour labels: for example, step $655$ corresponds to $H\approx300.209$ hours, labeled $300$ hours. Table~\ref{tab:data_ladder_budgets} lists the prescribed checkpoints, all evaluated with the protocol in Appendix~\ref{sec:impl}.

\begin{table}[t]
\centering
\caption{\textbf{Checkpoints in the balanced data ladder.} Each row is a cumulative budget along one training run per model configuration.}
\label{tab:data_ladder_budgets}
\small
\begin{tabular}{rr}
\toprule
Training budget (hours) & Optimizer steps \\
\midrule
$300$ & $655$ \\
$1{,}000$ & $2{,}182$ \\
$3{,}000$ & $6{,}545$ \\
$10{,}000$ & $21{,}818$ \\
$30{,}000$ & $65{,}454$ \\
\bottomrule
\end{tabular}
\end{table}

%% file: sections/g_appendix.tex
\section{Qualitative analysis}
\label{app:qual}

\subsection{Human samples}
\label{app:qual_human}

\begin{figure}[!htp]
  \centering
  \includegraphics[width=\linewidth]{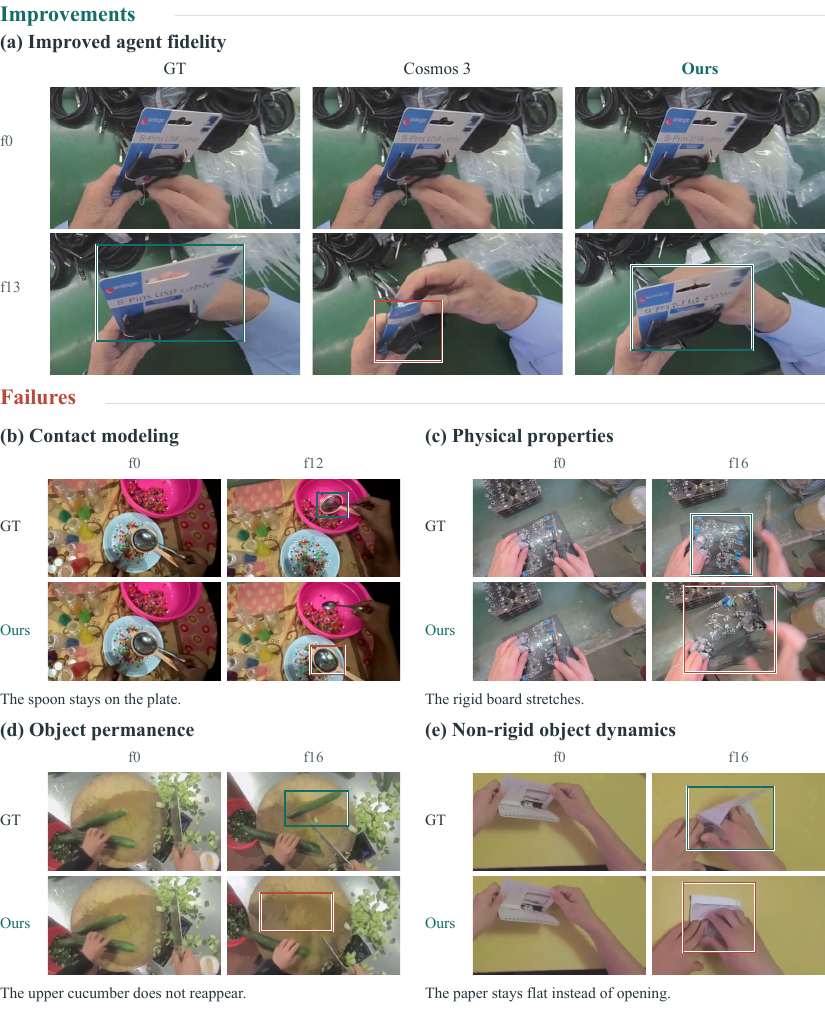}
  \caption{\textbf{Improvement and remaining failures in human manipulation.}
  (a)~Our model correctly captures the right hand's motion in the video, whereas vanilla Cosmos 3 fails.
  (b)~The spoon is not lifted.
  (c)~The rigid board stretches.
  (d)~The upper cucumber is missing after the hand moves away.
  (e)~The paper does not open.
  Frames and crops are matched across methods; f0 is the observed frame.
  Outlines mark the compared regions.  Best viewed on the \href{https://fidelity-gap.github.io/}{project website}.}
  \label{fig:qual_human}
\end{figure}

Figure~\ref{fig:qual_human} shows an example of improvement of our model over vanilla Cosmos~3 (top) and four remaining failure modes (bottom). Though our method improves the agent fidelity and some aspects of object interaction, as illustrated by the package handling sample, we still observe several failure modes. Specifically, contact modeling remains challenging, as illustrated by spoon example, where only one of the two objects is moved by the agent's right hand. The model also struggles with identifying physical properties of the objects from RGB frames, as shown on the example of the rigid board being interpreted as stretchable. In addition, object permanence and non-rigid object dynamics modeling remain a challenge, as can be seen from the disappearing cucumber sample and incorrect paper folding dynamics. More examples are provided on the \href{https://fidelity-gap.github.io/}{project website}, where they are also best viewed in a video form.

\subsection{Humanoid samples}
\label{app:qual_humanoid}

\begin{figure}[!htp]
  \centering
  \includegraphics[width=\linewidth]{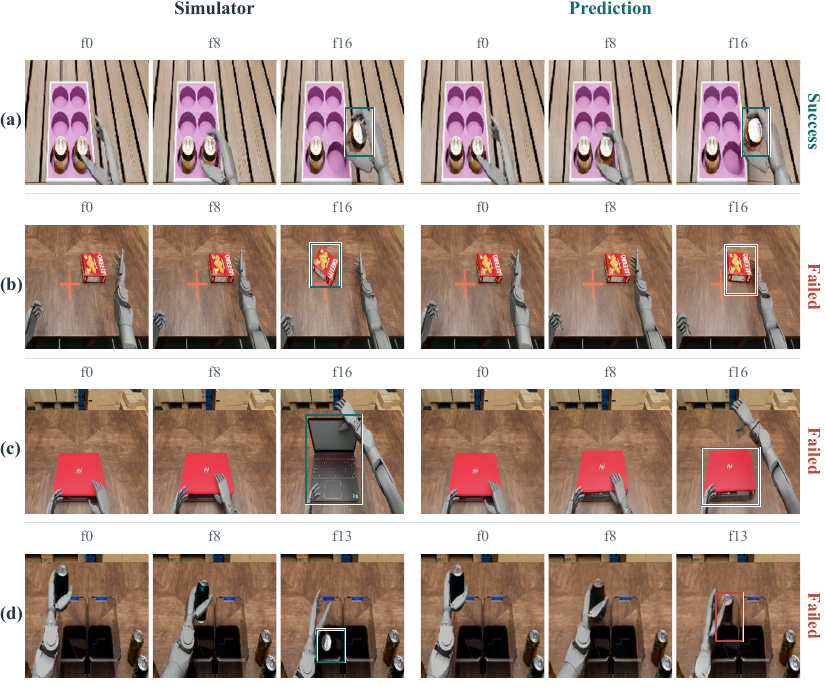}
  \caption{\textbf{One successful humanoid rollout and three object interaction failures.}
  Each row shows one example, with simulator's frames on the left and the corresponding predictions of our world model on the right.
  (a)~Both rollouts unload the can within the shown window.
  (b)~In the simulator the box turns, while in the world model it stays in place.
  (c)~In the simulator the laptop opens, while the world model leaves it closed.
  (d)~In the simulator the can drops in the bin, while in the world model's predictions it remains stuck near the bin's edge.
  Frame indices are aligned across each pair, with a common crop throughout.
  Success labels indicate whether the world model's prediction matches the simulator's object response. Best viewed on the \href{https://fidelity-gap.github.io/}{project website}.}
  \label{fig:qual_humanoid}
\end{figure}

Figure~\ref{fig:qual_humanoid} compares simulator rollouts with skeleton-conditioned predictions of our model. A successful unloading example is followed by three cases in which the object's response to agent's actions differs in the world model and in the simulator: a box does not turn under a push, a laptop does not open as the hand rises, and a can does not drop as the fingers open. Thus, accurate agent modeling does not always result in a faithful prediction of the manipulated object even in the simplified simulator environment. More examples are provided on the \href{https://fidelity-gap.github.io/}{project website}, where they are also best viewed in a video form.

%% file: sections/c_appendix.tex
\section{Additional Experiments}
\subsection{Skeleton Dropout Ablation}
\label{app:drop}

We test whether always supplying the projected skeleton limits object interaction learning by reducing the model's reliance on visual inference. We compare Cosmos~3 Nano with standard skeleton conditioning against a variant trained with skeleton dropout. For each training clip, we independently set the entire encoded skeleton sequence to zero with probability $0.25$, after the frozen VAE and before the learned projection. A single dropout decision applies to all frames and spatial locations in the clip. This intervention leaves the action tokens, observed RGB frame, and video targets unchanged. Both variants receive the full skeleton sequence during evaluation.

We compare checkpoints at 10k training hours, with the same pretrained initialization, training-data schedule, global batch size of 512, and optimizer settings. Both runs use standard denoising supervision without our dynamic-region modifications. 

\begin{table}[h]
\centering
\caption{\textbf{Skeleton dropout ablation.} Cosmos~3 Nano at 10,000 training hours. Dropout is applied only during training; both variants use full skeleton conditioning at evaluation. The results indicate that skeleton conditioning does not introduce shortcuts for learning object interaction dynamics.}
\label{tab:skeleton_dropout}
\begin{tabular}{lcc}
\toprule
Training condition & Agent SCS $\uparrow$ & Object SCS $\uparrow$ \\
\midrule
Skeleton conditioning & 0.779 & 0.513 \\
\quad + 25\% skeleton dropout & 0.778 & 0.509 \\
\bottomrule
\end{tabular}
\end{table}

As shown in Table~\ref{tab:skeleton_dropout}, the random dropout shows no meaningful effect on agent fidelity and marginally lowers object fidelity. At the matched 10k-hour budget, we therefore observe no scaling benefits from it, lifting the potential concerns of a learning shortcut introduced by skeleton conditioning.

\subsection{Perceptual scaling}
\label{app:perceptual_scaling}
\begin{figure}[htbp]
\centering
\includegraphics[width=\linewidth]{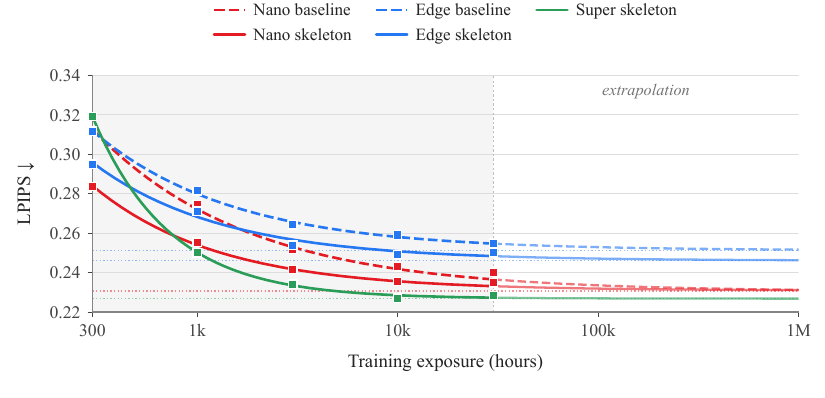}
\caption{\textbf{Perceptual quality across the data-scaling ladder.} LPIPS (lower is better) versus training exposure. Further scaling of ego-centric data is unlikely to improve the perceptual quality of the model's predictions either.}
\label{fig:lpips_scaling}
\end{figure}

Figure~\ref{fig:lpips_scaling} reports LPIPS (AlexNet) for the configurations in Figure~\ref{fig:data_scaling} on the validation set, averaged over the $16$ predicted frames and excluding the observed frame. We observe similar saturation behavior to that seen in the main paper on interaction fidelity metrics.

%% file: sections/d_appendix.tex
\section{Dynamic-Region Supervision Details}
\label{app:dynamic_region_details}

This appendix provides further details on the estimation of the dynamic-region map $M^{\mathrm{dyn}}$ and the two objectives it controls. We compute $M^{\mathrm{dyn}}$ by first measuring displacement in a common 3D reference frame, where camera motion is factored out and only genuine scene motion remains, and weighting that signal with a soft spatial prior around the projected hands, which
the skeleton conditioning already supplies at no additional cost. The result is a per-location score, high where something is both moving in the world and near the agent's hands, which we then use either to raise the noise level or to reweigh the loss.

\begin{figure}[t]
\centering
\includegraphics[width=\linewidth]{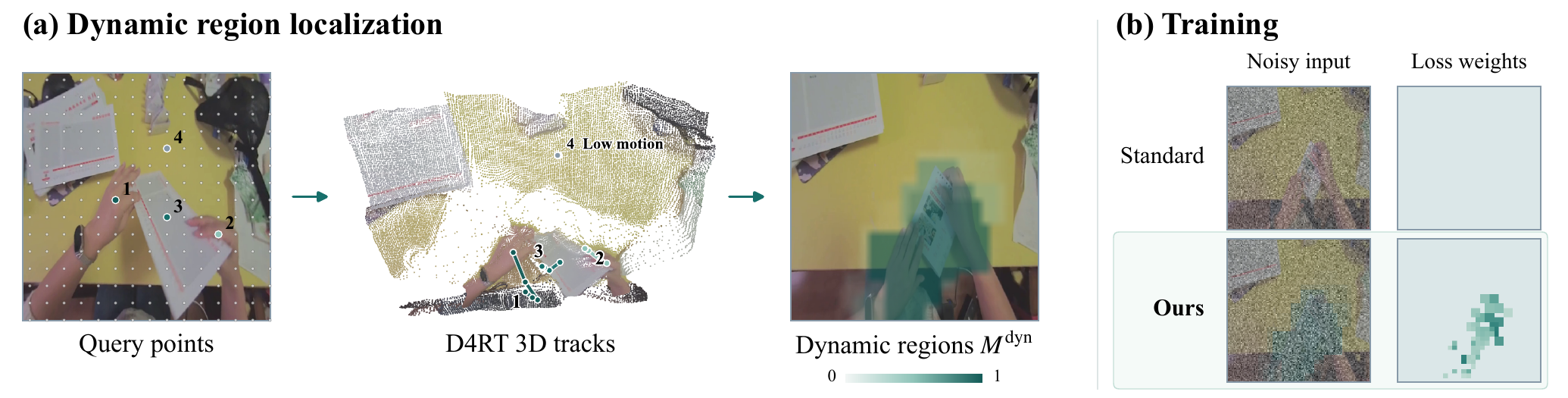}
\caption{\textbf{Dynamic-region supervision.} \textbf{(a)} D4RT tracks queried points in 3D to separate scene motion from camera motion. We combine 3D motion, tracking reliability, and a soft spatial prior around the hands to obtain the dynamic region map $M^{dyn}$. \textbf{(b)} Dynamic-region supervision combines dynamic noise scheduling
and loss reweighting in selected regions.}
\label{fig:object_supervision}
\end{figure}

\paragraph{Point tracks and motion scores.}
We initialize $256$ queries on a regular $16\times16$ grid in the first frame (shown in Figure~\ref{fig:object_supervision}, left). The video VAE encodes the initial observation as one latent frame and
temporally compresses the subsequent frames by a factor of four.
Thus, a 17-frame video produces $1+(17-1)/4=5$ latent frames.
We align tracking outputs to latent frame $k$ using video frame
$f(k)=4k$, for $k\in\{0,\ldots,4\}$, corresponding to video frames
$0,4,8,12,16$. For query $n$, D4RT provides an image position $\mathbf p_{kn}$ and a 3D displacement $\Delta X_{kn}$ from the source frame, measured in a common 3D reference. We multiply tracking confidence and visibility to obtain reliability $\gamma_{kn}\in[0,1]$. Queries are invalid if any position, confidence, visibility, or displacement is non-finite, the image position is out of bounds, or confidence is below $0.1$. For valid queries,
\begin{equation}
m_{kn}=\gamma_{kn}\min\!\left(\frac{\|\Delta X_{kn}\|_2}{\tau},1\right),
\qquad \tau=0.01.
\label{eq:dynamic_motion_score}
\end{equation}
Invalid queries receive zero score. The saturation scale $\tau$ uses the tracker's native 3D coordinate units.

\paragraph{Hand-proximity prior.}
Let $\mathcal H_{f(k)}$ contain the valid projected hand joints and bone segments in frame $f(k)$. We define
\begin{equation}
\begin{aligned}
d^{\mathrm{hand}}_{kn}&=\operatorname{dist}(\mathbf p_{kn},\mathcal H_{f(k)}),\\
\rho_{kn}&=\operatorname{sigmoid}\!\left(\frac{d^{\mathrm{hand}}_{kn}-8}{4}\right)
\operatorname{sigmoid}\!\left(\frac{48-d^{\mathrm{hand}}_{kn}}{4}\right).
\end{aligned}
\label{eq:dynamic_hand_proximity}
\end{equation}
Distances are measured in pixels on the $512\times512$ tracking canvas. This prior favours a soft band around the hand geometry, tapering near the skeleton primitives and beyond the hand neighbourhood. It is a spatial prior, not a contact label. We set $\rho_{kn}=0$ for non-finite positions or absent valid hand geometry; a bone segment is valid when both endpoints are valid and its length is nonzero.

\paragraph{Dynamic-region map.}
For an image of size $H_I\times W_I$ and latent grid of size $H_z\times W_z$, the continuous latent coordinates of a tracked point are
\begin{equation}
\bar{\mathbf p}_{kn}=\left(
\frac{W_z}{W_I}(p^x_{kn}+\tfrac12)-\tfrac12,\;
\frac{H_z}{H_I}(p^y_{kn}+\tfrac12)-\tfrac12\right).
\end{equation}
Its bilinear contribution to latent cell $\mathbf r=(r_x,r_y)$ is
\begin{equation}
\omega_{kn}(\mathbf r)=\kappa(r_x-\bar p^x_{kn})\kappa(r_y-\bar p^y_{kn}),
\qquad \kappa(a)=\max(0,1-|a|).
\end{equation}
Each query contributes to at most four cells; non-finite positions and out-of-grid contributions are discarded. We multiply the motion score by the hand prior, aggregate overlapping contributions by their maximum, and expand the sparse support:
\begin{equation}
\begin{aligned}
B_k(\mathbf r)&=\max_n\{\omega_{kn}(\mathbf r)m_{kn}\rho_{kn}\},\\
M_k^{\mathrm{dyn}}(\mathbf r)&=M_k^{\mathrm{pred}}(\mathbf r)
\operatorname{clip}_{[0,1]}\!\left(
\frac{\max_{\|\mathbf r'-\mathbf r\|_\infty\leq2}B_k(\mathbf r')}{0.1}
\right).
\end{aligned}
\label{eq:dynamic_region_fields}
\end{equation}
Cells without contributions have $B_k(\mathbf r)=0$. The neighbourhood maximum is $5\times5$ max pooling with zero padding. The prediction mask $M^{\mathrm{pred}}$ sets the map to zero on the conditioning frame. We write $i=(k,\mathbf r)$ and $\Omega=\{i:M_i^{\mathrm{pred}}=1\}$ for the predicted locations; all spatial weights are shared across latent channels.

\paragraph{Dynamic-region supervision.}
In Equation~\ref{eq:paired_regional_corruption},
$\lambda\in[0,1]$ scales the local noise increment $\delta_i$:
$\lambda=0$ gives $\delta_i=0$, while larger values increase
$\delta_i$ for fixed $\sigma$ and $M_i^{\mathrm{dyn}}>0$.
We use $\lambda=0.65$; for example, at $M_i^{\mathrm{dyn}}=1$
and $\sigma=0.5$, the increment is $\delta_i=0.325$, giving an
effective local noise proportion of $0.825$ while the model's
noise-level input remains $0.5$. For $i\in\Omega$,
$0<\sigma<1$, and $v_i=\epsilon_i-z_i$, the update can be written as
\begin{equation}
\begin{aligned}
\delta_i&=\lambda M_i^{\mathrm{dyn}}\min(\sigma,1-\sigma),\\
\widetilde z_{\sigma,i}&=(1-\sigma-\delta_i)z_i+(\sigma+\delta_i)\epsilon_i,\\
\widetilde v_i&=\left(1+\frac{\delta_i}{\sigma}\right)v_i.
\end{aligned}
\end{equation}
Because $M_i^{\mathrm{dyn}},\lambda\in[0,1]$, the local noise proportion satisfies
$\sigma\leq\sigma+\delta_i\leq1$. Input and target are modified together: the model still
receives the nominal level $\sigma$, and the target preserves the clean latent,
\begin{equation}
\widetilde z_{\sigma,i}-\sigma\widetilde v_i
=z_i+(\sigma+\delta_i)v_i-(\sigma+\delta_i)v_i=z_i.
\end{equation}
The conditioning frame is left unmodified and excluded from the loss.

\paragraph{Reweighting the loss.}
As an alternative to modifying the noise level, the same map can be used to reweight the loss within the dynamic regions. With gain $g\geq0$,
\begin{equation}
\begin{aligned}
w_i &= \frac{1+gM^{\mathrm{dyn}}_i}{\langle1+gM^{\mathrm{dyn}}\rangle_\Omega},\\
\mathcal{L}_{\mathrm{reg}}
&=\mathbb{E}_{z,\epsilon,\sigma}\!\left[
\sum_{i\in\Omega}w_i\,
\left\|v_\theta(\widetilde z_\sigma,\sigma;c,a_{0:T})_i-\widetilde v_i\right\|_2^2
\right],
\end{aligned}
\label{eq:regional_training_objective}
\end{equation}
where $\langle\cdot\rangle_\Omega$ is the mean over predicted locations within a clip, so that $\langle w\rangle_\Omega=1$ and emphasis is redistributed without changing the average weight. We use $g=2$. Locations with $M_i^{\mathrm{dyn}}=0$ retain a positive baseline weight. When the entire map is zero, $w_i=1$ and both operations reduce to standard training.

\paragraph{Random-region baseline.}
For the \emph{Dynamic noise (random)} baseline in Table~\ref{tab:ablation}, we replace the dynamic-region map with a randomly placed rectangle on the latent spatial grid.
For each clip, we sample its center uniformly over the grid and its aspect ratio in normalized coordinates as $\log r\sim\mathcal U(\log 0.5,\log 2)$.
With a fixed area fraction $a=0.15$, its width and height are
\begin{equation}
w=W_z\sqrt{ar},\qquad h=H_z\sqrt{a/r}.
\end{equation}
If part of the rectangle extends beyond the left or right edge
of the latent grid, that part reappears at the opposite edge;
the same rule applies to the top and bottom edges.
We treat each spatial location in the latent grid as a unit square.
Its mask value is the fraction of that square covered by the
rectangle: $1$ for full coverage, $0$ for no coverage, and a
fractional value for partial coverage.
This preserves a mean mask value of $0.15$ on each predicted frame, including for rectangles crossing a boundary.
We repeat the same spatial mask across all predicted latent frames and set it to zero on the conditioning frame.
Region selection is independent of image content, hand geometry, and tracking outputs. We apply dynamic noise scheduling with $\lambda=0.65$
and uniform loss weights ($g=0$).

%% file: sections/b_appendix.tex
\section{Humanoid Evaluation Details}
\label{app:humanoid}

\paragraph{Fine-tuning setup.}
The humanoid state space is 138-dimensional against the pre-trained checkpoints' 160, so we
re-initialize the action projection and leave the rest of the model unchanged. All three
variants are fine-tuned on ground-truth demonstrations from the same 11 tasks with the hyperparameters of Section~\ref{sec:impl}, with a global batch size of 64. For \textsc{Ours}, skeleton conditioning and dynamic-region supervision remain active during fine-tuning, with the skeleton obtained from the humanoid's forward kinematics.
 
\paragraph{Evaluation data and masks.}
SCS is measured on the benchmark's validation split. Hand masks come from SAM2~\citep{sam2} prompted at frame~0 with 18 forward-kinematics-projected links per hand and then tracked; object masks come from SAM3 with a per-task text prompt. All masks were manually verified. Success agreement is measured on a separate set of hand-curated windows, so the two measures agree on the ordering of the variants while being computed over different clips.
 
\paragraph{Checkpoint selection.}
Each variant is reported at its best checkpoint within a capped window rather than over its whole run, because the three converge at very different rates
(Table~\ref{tab:humanoid-iters}): both ego-centrically pre-trained variants are at their plateau by iteration 2{,}000, while \textsc{Cosmos~3} gains 0.19 hand SCS between 2{,}000 and 9{,}000 on trained tasks against 0.003 for \textsc{Ours}. Reading the comparison at a single early iteration
would therefore measure convergence rate rather than the effect of pre-training. We cap \textsc{Ours} and \textsc{+Human} at 3{,}000 and \textsc{Cosmos~3} at 9{,}000. Selecting on the clips we report is optimistic in absolute terms, but it is applied identically to all three variants; at fixed iterations the two pre-trained variants differ from their selected values by at most 0.005 SCS.

\begin{table}[t]
\centering
\caption{SCS at fixed iterations, averaged over all 11 tasks. The ego-centrically pre-trained
variants are converged throughout this range; \textsc{Cosmos~3} is not, which is why
Table~\ref{tab:humanoid-ablation} selects per variant rather than fixing a shared iteration.}
\label{tab:humanoid-iters}
\begin{tabular}{llccc}
\toprule
metric & variant & 2{,}000 & 3{,}000 & 9{,}000 \\
\midrule
\multirow{3}{*}{hand SCS}   & \textsc{Ours}      & 0.88 & 0.88 & \textbf{0.88} \\
                            & \textsc{+Human}    & 0.70 & 0.70 & 0.69 \\
                            & \textsc{Cosmos~3}  & 0.44 & 0.56 & 0.64 \\
\addlinespace
\multirow{3}{*}{object SCS} & \textsc{Ours}      & \textbf{0.77} & 0.77 & 0.76 \\
                            & \textsc{+Human}    & 0.70 & 0.71 & 0.72 \\
                            & \textsc{Cosmos~3}  & 0.56 & 0.67 & 0.67 \\
\bottomrule
\end{tabular}
\end{table}

\paragraph{Per-task results.}
Table~\ref{tab:humanoid-pertask} reports SCS for each of the 11 tasks and all three variants.
 
\begin{table}[t]
\centering
\caption{Per-task SCS on the humanoid benchmark, at the checkpoints reported in
Table~\ref{tab:humanoid-ablation}. }
\label{tab:humanoid-pertask}
\setlength{\tabcolsep}{5pt}
\begin{tabular}{@{}lccc@{\hskip 14pt}ccc@{}}
\toprule
 & \multicolumn{3}{c}{hand SCS} & \multicolumn{3}{c@{}}{object SCS} \\
\cmidrule(lr){2-4}\cmidrule(l){5-7}
task & \textsc{Cosmos~3} & \textsc{+Human} & \textsc{Ours}
     & \textsc{Cosmos~3} & \textsc{+Human} & \textsc{Ours} \\
\midrule
Close-Drawer                       & 0.60 & 0.65 & 0.88 & 0.92 & 0.86 & 0.97 \\
Flip-Mug                           & 0.69 & 0.75 & 0.90 & 0.70 & 0.77 & 0.79 \\
Insert-Cans                        & 0.59 & 0.64 & 0.86 & 0.60 & 0.61 & 0.62 \\
Open-Drawer                        & 0.53 & 0.65 & 0.86 & 0.58 & 0.49 & 0.62 \\
Pour-Balls                         & 0.71 & 0.77 & 0.89 & 0.75 & 0.78 & 0.81 \\
Push-Box                           & 0.65 & 0.79 & 0.89 & 0.81 & 0.86 & 0.91 \\
Stack-Can                          & 0.72 & 0.74 & 0.87 & 0.55 & 0.54 & 0.70 \\
Unload-Cans                        & 0.66 & 0.64 & 0.90 & 0.67 & 0.55 & 0.74 \\
Open-Laptop                        & 0.62 & 0.72 & 0.92 & 0.77 & 0.89 & 0.92 \\
Sort-Cans                          & 0.68 & 0.75 & 0.86 & 0.67 & 0.84 & 0.87 \\
Stack-Can-Into-Drawer              & 0.47 & 0.61 & 0.82 & 0.64 & 0.83 & 0.75 \\
\midrule
mean                               & 0.63 & 0.70 & 0.88 & 0.70 & 0.73 & 0.79 \\
\bottomrule
\end{tabular}
\end{table}

\paragraph{Success-agreement protocol.}
From each EgoVLA rollout we extract the 17 frames in which the task outcome is decided, e.g. the moment an object is released, a target is reached, or the simulator records success. We replay that clip's actions through each variant and render the prediction. A reviewer then records a success or fail verdict from the rendered clip, once for the simulator's rendering and once for each variant's prediction, applying the benchmark's criterion in both cases, and we report how often the two verdicts agree.

The verdict is therefore about the window, not the episode. This matters because a 17-frame window covers a single interaction event, so a failed episode can contain a cleanly executed sub-task and a successful one a failure. Using the simulator's episode-level flag as the reference would mislabel a substantial fraction of windows. We use that flag only to stratify which clips are reviewed, so the reviewed subset is not skewed toward whichever outcome dominates a task.

Seven of the eleven tasks have such a decisive moment: \textsc{Close-Drawer},
\textsc{Insert-Cans}, \textsc{Open-Laptop}, \textsc{Pour-Balls}, \textsc{Push-Box}, \textsc{Sort-Cans} and \textsc{Stack-Can}. In the other four the outcome unfolds over more frames than the clip contains like in 
\textsc{Open-Drawer} and \textsc{Unload-Cans}; or the success conditions defined in the simulator are too lenient like in \textsc{Flip-Mug}. Therefore we exclude these tasks from our evaluation. The review covers six clips per task, 42 in total, with one reviewer and one seed per variant (detailed breakdown provided in Table~\ref{tab:humanoid-parity-agree}).

\begin{table}[t]
\centering
\caption{Agreement with the simulator's rendering on the seven eligible tasks. \emph{Agreement}
is the fraction of clips on which the two verdicts coincide; recalls are conditioned on the
verdict for the simulator's rendering.}
\label{tab:humanoid-parity-agree}
\begin{tabular}{lcccc}
\toprule
variant & success & agreement & fail recall & success recall \\
\midrule
simulator rendering & 0.86 & ---           & ---  & --- \\
\textsc{Cosmos~3}   & 0.67 & 0.67          & 0.50 & 0.69 \\
\textsc{+Human}     & 0.67 & 0.71          & 0.67 & 0.72 \\
\textsc{Ours}       & 0.76 & \textbf{0.81} & 0.67 & \textbf{0.83} \\
\bottomrule
\end{tabular}
\end{table}

%% file: sections/f_appendix.tex
\section{Evaluation Details}
\label{sec:app_eval}
\begin{figure}[t]
  \centering
  \includegraphics[width=\linewidth]{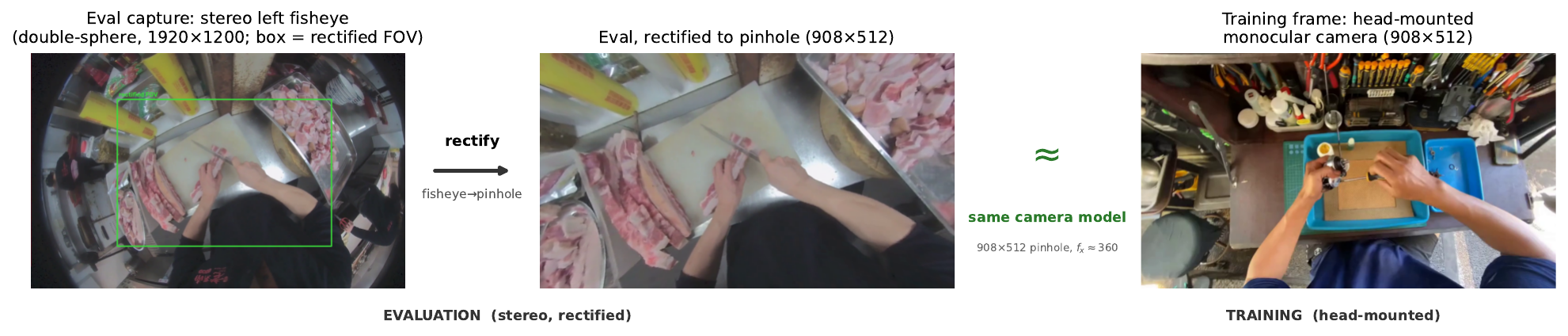}
  \caption{\textbf{Rectifying the evaluation set onto the training camera model.}
  \emph{Left:} the raw capture from the stereo rig's left camera, a wide-angle
  double-sphere fisheye ($1920{\times}1200$); the green box marks the region that
  maps into the rectified image. \emph{Middle:} that region undistorted to a
  $908{\times}512$ pinhole. \emph{Right:} a training frame, shown for comparison. }
  \label{fig:rectification}
\end{figure}

Our evaluation sequences are captured with a multi-camera stereo platform, whereas our training dataset is monocular. We use the stereo left camera for evaluation, however, it is delivered as a double-sphere fisheye. To match the training set sensor, for each frame we unproject the target $908{\times}512$ pinhole pixels to rays and reproject them through the double-sphere model, resampling the source
image (Fig.~\ref{fig:rectification}); the intrinsics are held fixed across all
clips. This yields a pinhole camera model consistent with the training data,
isolating content/appearance domain shift from any lens-model mismatch.

%% file: sections/e_appendix.tex
\section{Implementation Details}
\label{sec:impl}
We use $17$-frame RGB sequences at $10$\,fps and $512\times512$ resolution: one observed frame followed by $16$ predicted frames, spanning $1.6$\,s. We use AdamW with global batch size $512$, learning rates of $2\times10^{-4}$ for video-generation and skeleton-projection parameters and $10^{-3}$ for the action interface, $(\beta_1,\beta_2)=(0.9,0.99)$, $\epsilon=10^{-6}$, and zero weight decay. Learning rates are constant, with no warmup. Each configuration follows one progressive pass through the training manifest. The five data budgets are checkpoints along this trajectory, so optimization budget scales with data exposure. Appendix~\ref{app:data_scaling_ladder} specifies the sampling convention and checkpoint budgets.

Action normalization uses per-coordinate $1$st and $99$th percentiles, estimated separately for absolute states and consecutive-frame differences. We affinely map each percentile interval to $[-1,1]$ without clipping, using the absolute-state statistics for the initial action token and the delta statistics for subsequent tokens. These statistics remain fixed throughout training and evaluation.

We evaluate EMA checkpoints on the same $150$ held-out clips. Generation uses UniPC~\citep{zhao2023unipc} with $35$ sampling steps and a fixed classifier-free guidance scale of $5$. Agent and object SCS exclude the observed frame and average over future frames $1$--$16$. LPIPS also uses the same  $16$-frame window. We report the checkpoint at each prescribed budget, without selecting the best checkpoint by validation score.

When rendering agent's pose as a skeleton, we use the following hyper-parameters: finger hue is set to $\phi_f=\phi_{\mathrm{side}}+24f$, with $\phi_L=210$, $\phi_R=0$, and $f=0,\ldots,4$ from thumb to little finger; left/right arm hues are $180/25$ degrees. The parameters $(\alpha_{\mathrm{joint}},\alpha_{\mathrm{bone}},\beta)$ are $(0.6,0.9,0.65)$ for hands and $(0.55,0.75,0.55)$ for arms.